\documentclass[useAMS,usenatbib,referee]{biom}

\usepackage{float}
\usepackage{soul,color}
\usepackage[figuresright]{rotating}
\usepackage{amsmath}

\usepackage{comment}
\usepackage{diagbox}
\usepackage{amsfonts}
\usepackage{afterpage,lscape}
\usepackage{amssymb}
\usepackage{tabu}
\usepackage{caption}
\usepackage{mdwlist}
\usepackage{subfigure}
\usepackage{xcolor}
\usepackage{tabularx}
\usepackage{pifont}
\usepackage{mathtools}
\usepackage{booktabs}
\usepackage{threeparttable}
\usepackage{chngcntr}
\usepackage{setspace}
\usepackage{rotating}
\usepackage{graphicx}
\usepackage{titlesec}
\usepackage{multirow}
\titlespacing*{\section}{0pt}{0.5\baselineskip}{0.2\baselineskip}
\titlespacing*{\subsection}{0pt}{0.2\baselineskip}{0.2\baselineskip}

\allowdisplaybreaks
\title[snbClust]{A sparse negative binomial mixture model for clustering RNA-seq count data}

\author{Tanbin Rahman$^{1}$, 
Yujia Li$^{2}$, Tianzhou Ma$^{3}$, 	Lu Tang$^{2}$and George C. Tseng$^{2,*}$\email{ctseng@pitt.edu} \\
$^{1}$Department of Biostatistics, MD Anderson Cancer Center, Houston, TX 77030.\\
$^{2}$Department of Biostatistics, University of Pittsburgh, Pittsburgh, PA 15261.\\
$^{3}$Department of Epidemiology and Biostatistics, University of Maryland, Maryland 20742.
}

\begin{document}



\pagerange{\pageref{firstpage}--\pageref{lastpage}} 
\volume{64}
\pubyear{2008}
\artmonth{December}


\doi{10.1111/j.1541-0420.2005.00454.x}
\label{firstpage}

\begin{abstract}
Clustering with variable selection is a challenging yet critical task for modern small-n-large-p data. Existing methods based on sparse Gaussian mixture models or sparse $K$-means provide solutions to continuous data. With the prevalence of RNA-seq technology and lack of count data modeling for clustering, the current practice is to normalize count expression data into continuous measures and apply existing models with Gaussian assumption. In this paper, we develop a negative binomial mixture model with lasso or fused lasso gene regularization to cluster samples (small $n$) with high-dimensional gene features (large $p$). EM algorithm and Bayesian information criterion are used for inference and determining tuning parameters. The method is compared with existing methods using extensive simulations and two real transcriptomic applications in rat brain and breast cancer studies. The result shows superior performance of the proposed count data model in clustering accuracy, feature selection and biological interpretation in pathways.

\end{abstract}


\maketitle
\thanks{1}
\section{Introduction}
\label{sec:intro}

Cluster analysis is a powerful exploratory tool for high-dimensional data. In omics applications, many classical methods such as K-means clustering \citep{macqueen1967some}, hierarchical clustering \citep{eisen1998cluster}, self-organizing map (SOM) \citep{kohonen1998self} and model-based clustering \citep{fraley2002model} have been widely used. In transcriptomic data measured in microarray, for example, genes can be clustered into gene modules that suggest co-regulated or co-expressed genes with related biological function. In complex diseases, patients can be clustered to identify novel disease subtypes with distinct disease mechanism or drug responses, which often forms basis for personalized medicine and such sample clustering is the focus of this paper. When clustering such high-dimensional data, methods such as hierarchical clustering and SOM are heuristic in nature while model-based clustering assumes data to come from a mixture distribution of multiple clusters. Although the heuristic clustering algorithms are easy to implement and popular, they lack formal inference \citep{fraley2002model}. Model-based clustering, on the other hand, imposes distributional assumptions on data observations, and can allow rigorous inference, biological interpretation and prediction of future samples. In microarray, model-based clustering has been found with superior performance compared to heuristic methods such as hierarchical clustering or SOM \citep{thalamuthu2006evaluation}.

When clustering patients in omics data with thousands of genes, it is biologically reasonable to assume that only a small subset of genes (e.g. 50-200 genes) are cluster predictive. For this purpose, \cite{pan2007penalized}  proposed a Gaussian mixture model-based clustering with lasso penalty. \cite{witten2010framework} proposed a sparse $K$-means algorithm extended from $K$-means for feature selection. \cite{bouveyron2014model} and \cite{fop2018variable} provided thorough reviews of variable selection methods of clustering high-dimensional data. These methods can serve well for clustering continuous transcriptomic data from the gradually outdated microarray platforms. In the past ten years, the rapid development of RNA sequencing (RNA-seq) technology has revolutionized the transcriptomic research. Unlike the continuous florescent measurements from microarray, one important feature of RNA-seq is the (discrete) count-based data after alignment of millions of sequencing reads. In the literature, a common practice is to transform RNA-seq count data into continuous normalized values (e.g. TPM or CPM values) and directly apply methods that have been developed for microarray. This leads to significant loss of information, particularly for genes with lower counts. Methods directly modeling count data are expected to better fit the experimental data generation process, capture essential data characteristics, and thus perform better in clustering accuracy and gene selection.

 In the literature, \cite{si2013model} proposed a count-based model for clustering genes, where variable selection is not needed since $n$ is usually small compared to $p$.  \cite{witten2011classification} proposed to clustering RNA-sequencing data by hierarchical clustering where the dissimilarity matrix is calculated by Poisson assumption. This method neither considers potential overdispersion of count data nor achieves gene selection. In this paper, we focus on the problem of clustering samples with variable (gene) selection using transcriptomic data from RNA-seq. The data are count-based and usually contain $\sim$50-200 samples and $\sim$5,000-10,000 genes after proper gene filtering, which necessitates effective variable selection while performing clustering. Our approach directly deals with the count data using mixture of negative binomial model via GLM framework, without loss of information from transformation to continuous data. Further, we perform variable selection by lasso or fused lasso penalty to shrink the cluster specific means of each feature towards its global mean across all clusters. The paper is structured as follows. In Section \ref{sec:method}, we first summarize two existing methods, sparse Gaussian mixture clustering and sparse $K$-means (Section 2.1), and then propose the sparse negative binomial mixture clustering model (Section 2.2). Optimization for the penalized likelihood (Section 2.3), Bayesian information criterion (BIC) for model selection (Section 2.4) and performance benchmarks (Section 2.5) will be presented. Section \ref{sec:sim} will cover extensive simulations to benchmark and justify improved performance of the proposed methods. In Section \ref{sec:real}, two real applications using RNA-seq data from rat brain and breast cancer subtype studies will be evaluated to demonstrate improved clustering accuracy and gene selection. Section \ref{sec:discuss} contains final conclusion and discussion.

\section{Existing and proposed methods}
\label{sec:method}

We will briefly review two existing methods sparse Gaussian model-based clustering and sparse $K$-means in Section \ref{subsec:ctsmethod}.  To simplify discussion hereafter, we will abbreviate the \underline{s}parse \underline{G}aussian \underline{clust}ering model as ``sgClust" and abbreviate the \underline{s}parse \underline{$K$-means} method as ``sKmeans". We will then present our method \underline{s}parse \underline{n}egative \underline{b}inomial model-based \underline{clust}ering (snbClust) in Section 2.2. Section \ref{subsec:opt} discusses EM algorithm for optimizing the penalized likelihood function of ``snbClust". Section \ref{subsec:modelselect} and \ref{subsec:evaluation} will discuss Bayesian Information Criterion (BIC) for model selection and benchmarks for evaluation, respectively. In Section 2.6, we will illustrate the limitation of lasso penalty in Section 2.2, propose an alternative fused lasso penalty, evaluate the two and recommend to use the original lasso penalty for general practice.

Throughout the paper, we assume the raw sequencing reads from RNA-seq experiment are properly preprocessed, aligned and summarized. Denote by $y_{ij}$ the observed counts for gene $j$ ($1\leq j\leq G $) in sample $i$ ($1\leq i\leq n$). Our proposed snbClust model will utilize the count data as input. For the two existing methods, sgClust and sKmeans, Gaussian assumption is explicitly or implicitly assumed and only continuous input data are allowed. We will generate $\log$-transformed (base 10) CPM (Counts per Million) values using the edgeR package \citep{robinson2010edger}. The resulting log-CPM continuous values are denoted as $x_{ij}$ and are the input data for sgClust and sKmeans.


\subsection{Two existing methods using continuous input data}
\label{subsec:ctsmethod}

\subsubsection{sparse Gaussian model-based clustering model (sgClust)}
\label{subsubsec:sgClust}

\cite{pan2007penalized}  proposed a penalized likelihood approach by extending
from conventional Gaussian mixture model with a penalty term for feature
selection. By assuming zero mean for each gene vector, the penalty term is
simply the sum of $l_{1}$-norm of all cluster means in all genes. Specifically,
the likelihood to be maximized is
\begin{equation*}
\log L(\theta;x)=\sum\limits_{i=1}^{n}\log [\sum\limits_{k=1}^{K}p_{k}
f_{k}(x_i;\theta_k)]-\lambda b(\theta),
\end{equation*}
where $f_{k}(x_i;\theta_k)$ is the density function of multivariate normal distribution for cluster $k$ with cluster means and variances $\theta_k=\{\mu_{k},\Sigma_{k}\}$, $x_i=(x_{i1},\cdots,x_{iG})$, $p_k$ is the mixing probability of the $k$-th cluster and $b(\theta)=\sum\limits_{j=1}^{G}\sum\limits_{k=1}^{K}|\mu_{jk}|$  is the penalty term for regularization.
For simplicity, we note that this method assumes diagonal (i.e. independence across genes) and
equal covariance matrices across all clusters (i.e. $\Sigma_k=\operatorname{diag}\left(\sigma_{1}, \sigma_{2}, \ldots, \sigma_{G}\right), \forall k$). In real applications, each gene vector is standardized to zero mean before applying the method. Since no R package is available to the best of our knowledge, we wrote the R functions to carry out the algorithm and included it in our R package.

\subsubsection{Sparse $K$-means clustering (sKmeans)}
\label{subsubsec:sKmeans}
$K$-means clustering is a classical, efficient and effective clustering algorithm that seeks to minimize the within cluster sum-of-squares (WCSS). The method is related to Gaussian mixture model-based clustering with equal and spherical covariance matrices in each cluster \citep{tseng2007penalized}. In calculating distances for WCSS, traditional $K$-means adopts equal contribution from all gene features. In genomic applications, however, the input dataset contains thousands of genes and biologically only a small set of ``informative genes'' are relevant to sample clustering. \cite{witten2010framework} proposed a sparse $K$-means approach to allow feature selection and to improve clustering performance. While $K$-means minimizes the WCSS, sparse $K$-means equivalently seeks to maximize the between cluster sum of squares (BCSS) with gene-specific weight $w_j$ for gene $j$ and an $l_1$ lasso penalty on $w_j$. Specifically, sparse $K$-means seeks to optimize the following target function:
$$\max\sum\limits_{j=1}^{G} w_j\cdot BCSS_j=\sum\limits_{j=1}^{G} w_j\cdot (TSS_j-WCSS_j)$$
subject to $||w||^2\leq 1$, $||w||_1\leq s,$ and 
$w_j\geq 0, \forall j$. Here, $TSS_j= \frac{1}{n} \sum\limits_{i=1}^{n}\sum\limits_{i'=1}^{n}d_{j}(x_i,x_{i'})$ is the total sum-of-squares, $WCSS_j=\sum\limits_{k=1}^{K}\frac{1}{n_k}\sum\limits_{i,i'\in C_k}d_j(x_i,x_{i'})$ is the within cluster sum-of-squares for gene $j$, and $d_j(x_i, x_{i'})=(x_{ij}-x_{i'j})^2$. Note that $s$ is the tuning parameter to control feature selection (i.e. sparsity) and is chosen by gap statistic in the original paper. In this paper, the method is implemented using the R package ``sparcl''.

\subsection{Sparse negative binomial clustering with varying library size (snbClust)}
\label{subsec:snbClust}
Since RNA-seq experiment generates count data by nature, the common practice is to transform count data to continuous measures (e.g. $\log$CPM) and apply methods in Section \ref{subsec:ctsmethod}, thereby losing data information and reduce clustering performance. In the literature, negative binomial model has been widely used for RNA-seq differential expression analysis due to its better model fitness than Poisson model with an additional over-dispersion parameter. Assume, 
\begin{equation*}
y_{ij}|C_i=k \sim NB(\mu_{ijk},\phi_j);\hspace{0.2cm} \log(\mu_{ijk})= \log(s_{i})+\beta_{jk},
\end{equation*}
where $C_i$ is the cluster assignment for the $i$-th sample, $s_{i}$ is the normalization size factor of the $i$-th sample, a priori estimated by edgeR \citep{robinson2010edger} to control for the library size variation among samples,  $\beta_{jk}$ is the cluster mean of the $k$-th cluster for the $j$-th gene on the log scale after controlling for the library size variation and $\phi_j$ is the dispersion parameter for the $j$th gene. Here the negative binomial model is parameterized as $E(y_{ij}|C_i=k)=\mu_{ijk}$ and $Var(y_{ij}|C_i=k)=\mu_{ijk}+\frac{\mu^2_{ijk}}{\phi_j}$. Let  $\vec{y}_{i}=(y_{i1},y_{i2},\ldots,y_{iG})$ be the observed counts in sample $i$ with $G$ features. The penalized log-likelihood is given by,
\begin{equation}
\label{penalty}
\log L(\Theta_1)=\sum\limits_{i=1}^{n}\log[\sum\limits_{k=1}^{K}p_k f^{(nb)}_{k}(\vec{y}_i;s_{i}\exp (\vec{\beta}_k),\vec{\phi})]-\lambda h_0(\beta),
\end{equation}
where $\Theta_1=\{(p_k,\vec{\beta}_k),\vec{\phi};k=1,\ldots,K\}$ is the set of all unknown parameters, $f^{(nb)}_{k}(\vec{y}_i;s_i \exp(\vec{\beta}_k),\vec{\phi})$ is the density function of NB($s_i\exp(\vec{\beta}_{k}),\vec{\phi}$) with $\vec{\beta}_{k}=(\beta_{1k},\beta_{2k},\ldots \beta_{Gk}$) being the cluster means of cluster $k$, $\vec{\phi}=(\phi_1,\ldots,\phi_G)$ is the vector of gene-specific dispersion parameters and $p_{k}$ is the probability of belonging to cluster $k$. In the penalty term, $\lambda$ is the tuning parameter and $h_0(\beta) = \sum\limits_{k=1}^{K}\sum\limits_{j=1}^{G}|\beta_{jk}- \beta_{j}^{*}|$ with $\beta_{j}^{*}$ being the MLE of global mean of $j$-th gene on a log-scale assuming no cluster effect  after controlling for the library size variation (see section \ref{subsec:opt} for estimate of $\beta^*_j$). The formulation is similar to sgClust in Section 2.1.1, but we note that unlike the Gaussian model in sgClust in Section \ref{subsubsec:sgClust}, the count data can not be standardized in each gene row. The subtraction of overall global cluster mean $\beta^*_j$ for each gene $j$ in $h_0(\beta)$ is necessary. Maximization of the above likelihood can be achieved by using EM algorithm \citep{dempster1977maximum}. Here, we introduce a latent variable $z_{ik}=I{\{i\in C_k\}}$ as the indicator function of cluster assignment for sample $i$ to be assigned to cluster $k$ and the problem becomes maximizing the following complete penalized log-likelihood:
\begin{equation}
\label{eq5}
\log L_{c}(\Theta_2)=\sum\limits_{i=1}^{n}\sum\limits_{k=1}^{K} z_{ik}[\log (p_k)+\sum\limits_{j=1}^{G}\log (f^{(nb)}_{k}(y_{ij};\exp (\log(s_i)+\beta_{jk}),\phi_j))]-\lambda h_0(\beta),
\end{equation}
where $\Theta_2=\{(p_k, \vec{\beta}_k,\vec{z}_k) ; k=1,\ldots,K) \}$ and $\vec{z}_k=(z_{1k}, \ldots, z_{nk})$.
Details of optimization will be illustrated in the next subsection.

\noindent Remark: Gene selection result from the penalty $h_0(\beta) = \sum\limits_{k=1}^{K}\sum\limits_{j=1}^{G}|\beta_{jk}- \beta_{j}^{*}|$ in snbClust will differ from $b(\theta)=\sum\limits_{j=1}^{G}\sum\limits_{k=1}^{K}|\mu_{jk}|$ in sgClust in that $b(\theta)$ can select cluster-specific genes such as $(\mu_{j1},\cdots,\mu_{jK})=(3.74,0,\cdots,0)$ while $h_0(\beta)$ almost surely cannot achieve that. In Section 2.6, we will replace $\lambda h_0(\beta)$ and propose an alternative fused lasso penalty $h_1(\beta,\lambda)$ to alleviate the problem and perform simulation to evaluate the pros and cons.

\subsection{Optimization using EM algorithm}
\label{subsec:opt}
Expectation-Maximization (EM) algorithm is a method iterating between an expectation and a maximization step to find the maximum likelihood estimates of parameters in a model with unobserved latent variables (e.g. a mixture model with unknown cluster assignments in our case) . In the literature,  \cite{mclachlan1997algorithm} discussed the estimation of mixture of generalized linear models using  iteratively reweighted least square algorithm. \cite{friedman2010regularization} proposed the estimation of generalized linear model with convex penalties for variable selection using coordinate descent algorithm. For the estimation of snbClust model, we combined the above two ideas to derive a new EM algorithm to estimate the parameters in a mixture of generalized linear model with convex penalties in Equation \ref{eq5}. For the gene-specific dispersion parameters $\phi_j$'s, we estimated a priori by edgeR package and plugged into the model. For simplicity, $\vec{\phi}$ will be ignored as we introduce the algorithms below.




   


 We first pre-estimate $\beta_j^*$ (i.e. the global mean of non-informative feature $j$) and considered it known during the EM algorithm. $\beta_j^*$ is estimated by maximizing the following likelihood using  iteratively reweighted least square (IRLS) algorithm,

\begin{equation*}
 \sum\limits_{i=1}^{n}\sum\limits_{j=1}^{G} \log f(y_{ij}; \exp(\log(s_i)+\beta_{j}))
\end{equation*}
Once the vector $\beta^*_j$ is estimated, we carry out the EM algorithm as follows.
 The E-step yields:
\begin{eqnarray*}
 & Q(\Theta_2;\Theta_2^{(m)}) = E_{\Theta_2^{(m)}}(\log L_{c,\Theta_2} |Y) 
=\sum\limits_{i=1}^{n}\sum\limits_{k=1}^{K} z^{(m)}_{ik} [\log p_k+ \nonumber  \\  
&  \sum\limits_{j=1}^{G} \log f(y_{ij}; \exp(\log(s_i)+\beta_{jk}))]  - \lambda \sum_{j}^{G}\sum_{k}^{K}|\beta_{jk}-\beta_{j}^*|,
\end{eqnarray*}

\begin{equation*}
z_{ik}^{(m)}=\frac{p_k^{(m)}\prod\limits_{j=1}^{G}f^{(nb)}_{k}(y_{ij};\exp(\log(s_i)+\beta_{jk}^{(m)}))}{\sum\limits_{l=1}^{K}p_l^{(m)}\prod\limits_{j=1}^{G} f_l^{(nb)}(y_{ij};\exp(\log(s_i)+\beta_{jl}^{(m)}))}
\end{equation*}

In the M-step, the updating function of $p$ is given by,

\begin{equation*}
p_k=\sum\limits_{i=1}^{n}z_{ik}/n
\end{equation*}

The updating function of $\beta$ cannot be easily derived by maximizing the above Q function. We can solve it by using IRLS algorithm, a similar idea recently applied in \cite{wang2016penalized} under a regression setting. Suppose $t$ is the current iteration of IRLS, we will repeat the following four steps until convergence and return the final estimates of $\beta_{jk}$ as $\beta_{jk}^{(m+1)}$:  

\begin{enumerate}  
\item Calculate $w_{ijk}^{(t+1)} = \mu_{ijk}^{(t)}/( 1 + \phi_j^{-1} \mu_{ijk}^{(t)})  $
\item Update $\tau_{ijk}^{(t+1)} = \log(s_i) + \beta^{(t)}_{jk} + (y_{ij} - \mu_{ijk}^{(t)} )/\mu_{ijk}^{(t)}$
\item Solve $\beta_{jk}^{(t+1)}=$ argmin  $\frac{1}{2} \sum_i z_{ik}^{(m)} w_{ijk}^{(t+1)} (\tau_{ijk}^{(t+1)} -\log(s_i)-\beta_{jk})^2+ \lambda |\beta_{jk}-\beta_{j}^*| $
\item Update $ \mu_{ijk}^{(t+1)} = \exp(\beta_{jk}^{(t+1)} + \log(s_i)) $
\end{enumerate} 

The solution in step 3 is given by:
   
\begin{equation*}
\beta_{jk}^{(t+1)}=\beta_{j}^{*} +\text{sign}(\tilde{\beta}_{jk}-\beta_{j}^{*})[|\frac{ \sum_i z_{ik}^{(m)} w_{ijk}^{(t+1)} ( \tau_{ijk}^{(t+1)} - \log(s_i)) - \lambda\text{ sign}(\tilde{\beta}_{jk}-\beta_{j}^{*})}{\sum_i z_{ik}^{(m)} w_{ijk}^{(t+1)} }|-|\beta_{j}^{*}|]_{+}
\end{equation*}

where $\tilde{\beta}_{jk} = \sum\limits_{i=1}^{n} z_{ik}^{(m)} w_{ijk}^{(t+1)} ( \tau_{ijk}^{(t+1)} - \log(s_i)) / \sum\limits_{i=1}^{n} z_{ik}^{(m)} w_{ijk}^{(t+1)} $ is the estimate of $\beta_{jk}$ without penalization and $f_{+}$ is the soft-thresholding function which takes the value $f$ if $f_{+} > 0$
and 0 otherwise. 

Once we obtain the estimates $\beta_{jk}^{(m+1)}$ from the IRLS algorithm, we can continue to iteratively carry out E step and M step until convergence to obtain the final maximum penalized likelihood estimate (MPLE).

\subsection{Model selection}
\label{subsec:modelselect}

For snbClust, sgClust and sKmeans, number of clusters $K$ and the penalty tuning parameter $\lambda$ must be pre-estimated. How to simultaneously estimate $K$ and $\lambda$ is an understudied problem in the field and is out of the scope of this paper. Here following \cite{witten2010framework}, we assume $K$ has been estimated by other existing methods \citep{tibshirani2001estimating,sugar2003finding,tibshirani2005cluster} and then propose a model selection criterion to determine $\lambda$. BIC criterion \citep{schwarz1978estimating} is a popular method to determine the tuning parameter by minimizing the criterion. A modified version of the BIC was introduced by \citet{pan2007penalized} for the sgClust model. Here, we propose a similar BIC approach for estimating $\lambda$:
\begin{equation}
BIC=-2\log L(\hat{\Theta}_1)+\log (n)d_e, 
\end{equation}
where $\hat{\Theta}_1$ is the MPLE calculated from Section 2.3 given $K$ and $d_e=(K-1)+KG-q$ is the effective number of parameters. In determining $d_e$, the first term $K-1$ refers to the number of parameters in the mixing probabilities with constraint $\sum p_{k}=1$, the second term $KG$ is the number of parameters in cluster means. Finally, $q$ refers to the number of estimates (among the $K\cdot G$ cluster mean parameters) which are shrunken to the global mean. The dispersion parameters are pre-estimated, therefore they are considered known and thus not included in the BIC criterion.

For the snbClust and sgClust model, BIC criterion is used for selecting $\lambda$.  As for sKmeans, gap statistic was proposed in the original paper and software package `sparcl' is used for model selection. 

\subsection{Benchmarks for evaluation}
\label{subsec:evaluation}

In the high-dimensional clustering problem we consider here, the clustering performance is first benchmarked by the clustering accuracy using adjusted Rand index (ARI)\citep{hubert1985comparing} when the true cluster labels are known in simulations and real applications. We next consider performance on feature (variable) selection. In simulation, since the true cluster-predictive features are known, we use receiver operating characteristic (ROC) curve and its area under curve (AUC) for evaluation. In real data, the true cluster-predictive features are unknown. We perform pathway enrichment analysis using Fisher's exact test under different degrees of sparsity to evaluate statistical significance of biological annotation on selected features.

\subsection{Alternative fused lasso penalty for variable selection}
\label{subsec:Fusedlasso}
In Equation \ref{penalty},  lasso penalty $h_0(\beta) = \sum\limits_{k=1}^{K}\sum\limits_{j=1}^{G}|\beta_{jk}- \beta_{j}^{*}|$ is used to shrink the $\beta_{jk}$ towards $\beta_{j}^{*}$ for every feature $j$. The natural downside of lasso penalty is that grouping can only occur at $\beta_j^*$ but not elsewhere. When the penalty is not strong enough to shrink all $\beta_{jk}$ to $\beta_{j}^{*}$ for feature $j$, all $\beta_{jk}$'s tend to obtain distinct values. In other words, the lasso penalty cannot conclude for cluster-specific genes such as $(\hat\beta_{j1}, \hat\beta_{j2}, \hat\beta_{j3}, \hat\beta_{j4})=(5.96, 2.22, 2.22, 2.22)$ but instead will generate estimates like $(\hat\beta_{j1}, \hat\beta_{j2}, \hat\beta_{j3}, \hat\beta_{j4})=(5.96, 2.28, 2.24, 2.15)$. To accommodate this issue, we alternatively apply a fused lasso penalty $h_1(\beta, \lambda) = \sum_{j=1}^{G} \sum_{1 \le k < k' \le K} \rho (|\beta_{jk} - \beta_{jk'}|, \lambda)$, where $\rho(x, \lambda) \ge 0$ is a penalty function defined on ${\mathbb{R}^+_0}$, so that the distance between each pair of $\beta_{jk}$ and $\beta_{jk'}$ can be shrunken toward zero. In practice, we consider a nonconvex MCP penalty \citep{zhang2010nearly} ($P_{\gamma}(x ; \lambda)=\lambda|x|-\frac{x^{2}}{2 \gamma}$ if $|x| \leq \gamma \lambda$ and $P_{\gamma}(x ; \lambda)=\frac{1}{2} \gamma \lambda^{2}$ otherwise). In Web Appendix A, we develop an alternating direction method of multipliers (ADMM) \citep{ma2016exploration} method and embed it into IRLS and EM steps to solve the optimization. Based on the simulation results in Web Appendix B and C, the fused lasso penalty unfortunately only provides marginal benefit in limited scenarios, but the method is computationally more demanding and the result is more sensitive to the tuning parameter. For example, fused lasso penalty needs about 3 hours to finish simulation 2 (See simulation section for more detail) while lasso penalty needs 1.37 minutes on average. In general application, we will recommend to use lasso penalty $h_0(\beta) = \sum\limits_{k=1}^{K}\sum\limits_{j=1}^{G}|\beta_{jk}- \beta_{j}^{*}|$ and we will stick to this formulation hereafter. Details of the algorithm, simulation and results of fused lasso penalty are shown in the supplement materials.

\section{Simulation}
\label{sec:sim}

In this section, we conduct three simulations to show the advantages of snbClust while compare it to sKmeans and sgClust. In simulation 1, we assume that all genes are informative and all samples have equal library sizes. No variable selection is performed so we only assess the clustering performance. In simulation 2, we assume only a proportion of genes are informative and assess both the clustering accuracy and variable selection performance.  In simulation 3, we perform additional sensitivity analysis by simulating gene-gene dependency structure to examine whether the performance would be affected and whether the gene independence assumption is valid in general. We repeat 100 times for each simulation and evaluate the averaged results.

To mimic real data structure, we extract the main characteristics of The Cancer Genome Atlas (TCGA) breast cancer RNA-seq data, which is also used in the second real data example in Section \ref{sec:real2}, to perform the simulation. The dataset contains 610 female patients. We first compute the mean counts of each gene over all samples and obtaine an empirical distribution of mean counts, which will be used for obtaining baseline expression levels in all three simulations. Since RNA-seq data are usually skewed with many highly expressed house-keeping genes which are irrelevant to cluster analysis, we exclude the top 30\% mean counts when forming the empirical distribution. In addition, we also pre-estimate the gene-specific dispersion parameter $\vec{\phi}$ from the data using edgeR \citep{robinson2010edger} and use it for simulation.


\subsection{Simulation settings}

\noindent\underline{Simulation 1: no feature selection and equal library size}

\begin{enumerate}
\item[1.] Sample the baseline expression level of $G=150$ independent genes $\mu_j$ ($1\le j\le 150$) from the empirical distribution of mean counts constructed above.
\item[2.] Use $\delta_{jk} \in \{-1,0,1\}$ to represent the pattern of gene $j$ ($1\le j\le 150$) in cluster $k$ ($1\le k\le 3$), with $1$ indicating the gene is up-regulated in this cluster relative to baseline, $-1$ indicating down-regulation and $0$ indicating no difference. Assume there exists three gene patterns:  $(\delta_{j1},\delta_{j2},\delta_{j3})$ = $(-1,0,1)$ for $1\leq j\leq 50$, $(\delta_{j1},\delta_{j2},\delta_{j3})$ = $(0,1,1)$ for $51\leq j\leq 100$, and $(\delta_{j1},\delta_{j2},\delta_{j3})$ = $(1,-1,0)$ for $101\leq j\leq 150$.
\item[3.] Sample the log2 fold change (effect size) parameter $\Delta_{j}$ for gene $j$ ($1\le j\le 150$) and cluster $k$ ($1\le k\le 3$) from a truncated normal distribution $TN(\gamma, 1, \gamma/2, \infty)$ with mean $\gamma$, standard deviation 1 and $\gamma$/2 the lower truncation of the distribution (i.e. the minimal effect size). We vary the value of $\gamma\in \{0.4,, 0.6, 0.8, 1, 1.2 \}$ for a thorough comparison with the other methods.   
\item[4.] Denote class label $C_i=k$ for $1+(k-1)\cdot 15\leq i\leq k\cdot 15$ (i.e. 15 samples per cluster and 45 samples in total). Sample the count data by $y_{ij}|C_i=k \sim NB(\mu_j \times 2^{\Delta_{j} \times \delta_{jk}}; \phi )$ for gene $j$ ($1\le j\le 150$) and sample $i$ ($1\le i\le 45$) in cluster $k$ ($1\le k\le 3$). 

\end{enumerate}

\noindent\underline{Simulation 2: with feature selection}

\begin{enumerate}
\item[1.] As in simulation 1, sample the baseline expression level of $G=1000$ independent genes $\mu_j$ ($1\le j\le 1000$) from the empirical distribution of mean counts.
\item[2.] Similar to simulation 1, assume 150 genes are informative and there exist three gene patterns for these informative genes (50 genes in each): $(\delta_{j1},\delta_{j2},\delta_{j3})$ = $(-1,0,1)$, $(0,1,1)$ or $(1,-1,0)$. For non-informative genes, the pattern is $(0,0,0)$. 
\item[3.] Sample the log2 fold change (effect size) parameters $\Delta_{j}$ for gene $j$ ($1\le j\le 1000$) and cluster $k$ ($1\le k\le 3$) from a truncated normal distribution $TN(\gamma , 1, \gamma/2, \infty)$. Here, $\gamma \in \{0.60, 0.80, 1, 1.2\}$.
\item[4.] Sample the library size scaling factor $a_{i}$ from Unif (LB,UB) for each sample $i$ ($1\le i\le 45$), where LB and UB indicate the lower and upper bounds of the uniform distribution. We choose (LB,UB) to be $(0.9,1.1)$  and $(0.7,1.30)$ to compare snbClust to the other methods.  
\item[5.] Sample the count data by $y_{ij}|C_i=k \sim NB(a_i \mu_j \times 2^{\Delta_{j} \times \delta_{jk}}; \phi )$ for gene $j$ ($1\le j\le 1000$) and sample $i$ ($1\le i\le 45$) in cluster $k$ ($1\le k\le 3$).
\end{enumerate}

\noindent\underline{Simulation 3: sensitivity analysis under gene dependency}

\begin{enumerate}
\item [1.] For a total of $G=1000$ genes, assume 150 genes are informative and there exist three gene patterns for these informative genes (50 genes in each):  $(\delta_{j1},\delta_{j2},\delta_{j3})$ = $(-1,0,1)$, $(0,1,1)$ or $(1,-1,0)$. For non-informative genes, the pattern is $(0,0,0)$. 
\item [2.] Sample the log$_2$ fold change (effect size) parameters $\Delta_{j}$ for gene $j$ ($1\le j\le 1000$) and cluster $k$ ($1\le k\le 3$) from a truncated normal distribution $TN(0.5, 1, 0.25, \infty)$. 
\item [3.] Sample the baseline expression level $\mu_{j}$ ($1\le j\le 1000$) from the empirical distribution of mean counts. For each gene $j$ in cluster $k$, obtain $\theta_{jk} = \log_2(\mu_{j}) + \Delta_{j}\times \delta_{jk}$.
\item [4.] For each gene pattern, sample five gene modules, so there are a total of $M=15$ modules with $d = 10$ genes in each module ($1\le m\le 15$) for informative genes. 
\item [5.] Sample the covariance matrix $\Sigma_{mk}$ for genes in module $m$ ($1\le m\le 15$), cluster $k$ ($1\le k\le 3$). First sample $\Sigma'_{mk} \sim W^{-1}(\vec{\psi}, 60)$, where $\vec{\psi} = (1-\alpha)I_{d \times d} + \alpha J_{d \times d}$ and $\alpha \in \{0,0.25,0.5\}$ controls the correlation. $\Sigma_{mk}$ is calculated by standardizing $\Sigma'_{mk}$ so that the diagonal elements are all 1's. Here, $I_{d\times d}$ is an identity matrix of dimension $d \times d$ and $J_{d}$ is a matrix of 1 with dimension $d \times d$.
\item[6.] Sample the expression levels of all genes in each module $m$ as $(\beta_{i,(m-1)d+1},\beta_{i,(m-1)d+2} \ldots, \beta_{i,md})|C_i=k)$ $\sim MVN (( \theta_{(m-1)d+1, k}, \theta_{(m-1)d+2, k}\ldots, \theta_{md,k})^{T},\Sigma_{mk})$ for sample $i$ $(i\leq i \leq 45)$ in cluster $k$ $(1 \leq K \leq 3)$.
\item[7.] Sample the library size scaling factor $a_i$ from Unif(0.9,1.1) for each sample $i$ $(1 \leq i \leq 45)$.
\item[8.] Sample $y_{ij} \sim NB(a_i 2^{\beta_{ij}},\phi)$ for $1 \leq j \leq 150$ and sample  $i$ $(1 \leq i \leq 45).$ For $151 \leq j \leq 1000$, $y_{ij} \sim NB(a_i 2^{\mu_j},\phi)$.
\end{enumerate}

\subsection{Simulation results}

\begin{figure}
\centering
  \centerline{\includegraphics[width=7in]{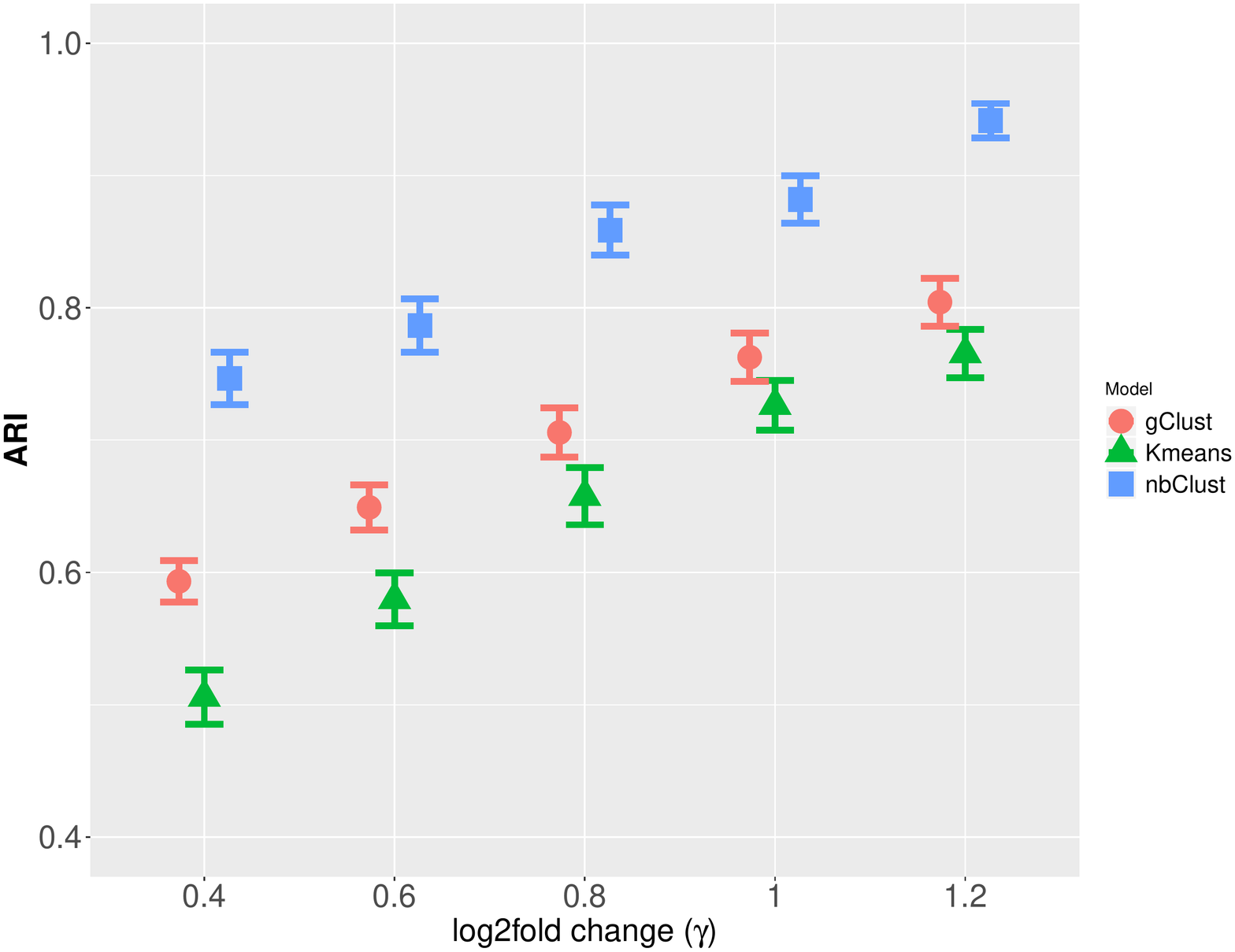}}
  \caption{ARI by effect size $\gamma$ for simulation scheme 1 when no feature selection is needed.}
\label{f:1}
\end{figure}

Figure \ref{f:1} shows the mean and standard error of ARI values over 100 replications for the three methods in Simulation 1. Here, the purpose is to evaluate whether using negative binomial distribution to model the count data outperforms other Gaussian-based methods in a simple situation. We consider all the genes to be informative; therefore, only clustering performance in terms of ARI is assessed in this case. For simplicity, we only consider 150 genes and the library size to be constant over all the samples.  Compared to Kmeans and gClust methods, nbClust has better clustering performance (larger ARI) and the advantage is consistent as we vary the minimal effect size $\gamma$. 
In Simulation 2, we evaluate how the performance varies when there are non-informative genes as well as varying effect size $\gamma$. The clustering performance is measured using the ARI as before while the variable selection is assessed using the AUC value. The result for this simulation scheme is summarized in Figure 2. Figure 2(A) shows the comparison of performance between the three methods when the variation of library size is moderate (normalization size factor varies from 0.90 to 1.10). The ARI value of snbClust is much higher on average compared to both sKmeans and sgClust. The variable selection performance in terms of AUC in Figure 2(B) is also higher for snbClust compared to the other two methods. When the signal strength $\gamma$ increases, we observe improved performance for ARI and AUC as expected. A similar trend is observed in the presence of high level of library size variation (normalization size factor varies from 0.70 to 1.30) shown in Figure 2(C)(D). 


    
    

\begin{figure}
\centering
  \centerline{\includegraphics[width=7in]{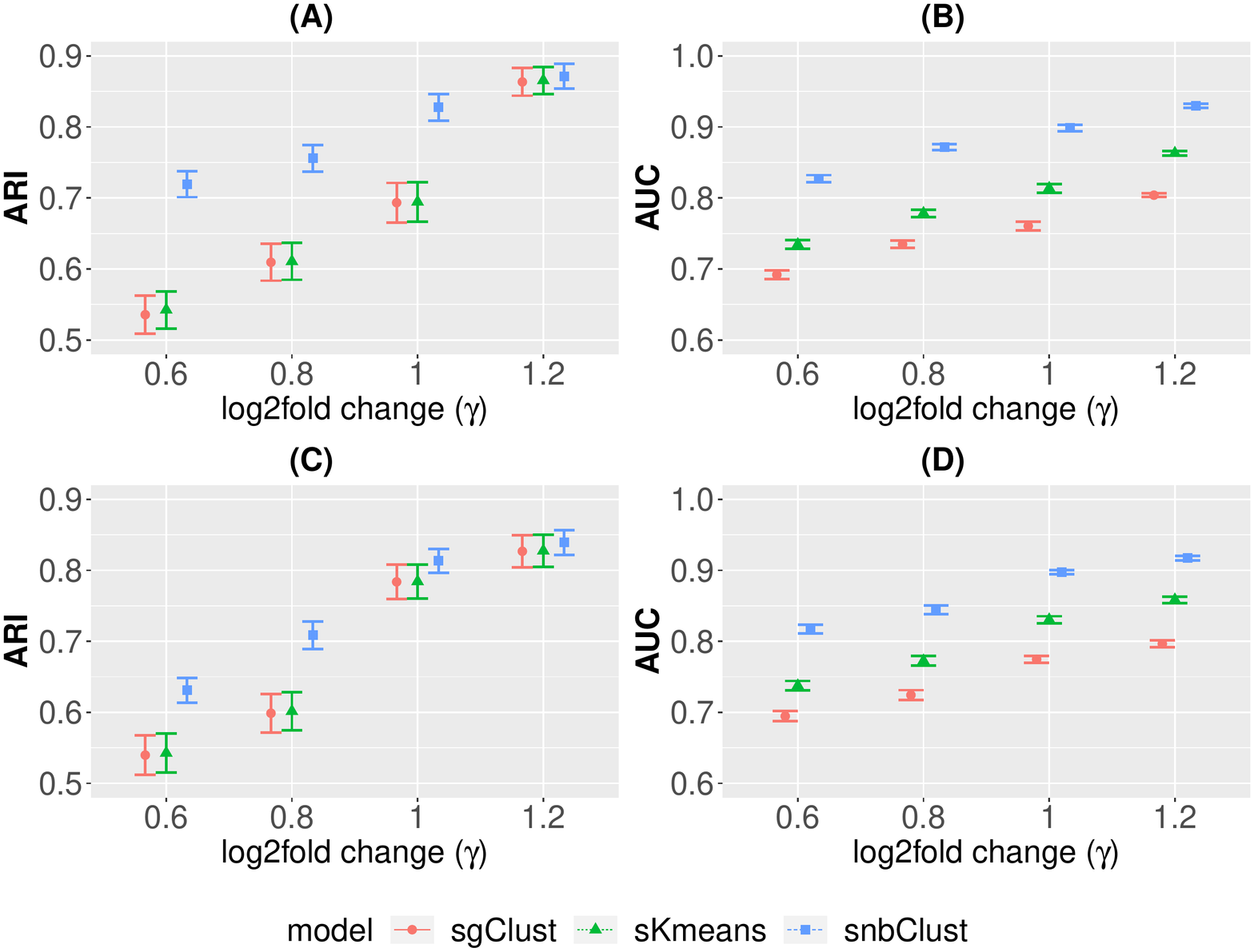}}
  \caption{Clustering accuracy by ARI and feature selection accuracy by AUC for Simulation scheme 2. Figure 2(A)(B) is the result of low library size variation (normalization size factor varies from 0.90 to 1.10); Figure 2(C)(D) is the result of high library size variation (normalization size factor varies from 0.70 to 1.30.}
\label{f:2}
\end{figure}

Table 1 shows the results in Simulation 3 for varying gene dependnce correlation $\alpha$. As we can see, the performance of snbClust remains relatively stable even when $\alpha$ increases up to 0.75, partially justifying the gene-gene independence assumption in our model. Intuitively, in high dimensional space, data points are much better separated and ignoring gene dependence structure may less impact the clustering performance, similar to the phenomenon of blessings of dimensionality described in \cite{donoho2000high}.


\begin{table}
	\centering

	\begin{tabular}{|c|c|c|c|}
		\hline
		Correlation & Model & ARI  & AUC \\ 
		\hline
		& snbClust & 0.822(0.01) & 0.892(0.005) \\ 
		0.25& sgClust & 0.767(0.026)  & 0.758(0.006) \\ 
		& sKmeans & 0.770(0.025) & 0.819(0.006)  \\ 
		\hline
		& snbClust & 0.811(0.018) & 0.896(0.004) \\ 
		0.50 & sgClust & 0.769(0.024) & 0.761(0.005)\\ 
		& sKmeans & 0.768(0.023)  & 0.826(0.005) \\ 
		\hline
		& snbClust& 0.831(0.017)  & 0.895 (0.003) \\ 
		0.75 & sgClust & 0.768(0.025)  & 0.772(0.005)  \\ 
		& sKmeans & 0.770(0.025) & 0.823(0.005)  \\ 
		\hline
	\end{tabular}
\caption{ARI and AUC performance when gene-gene correlation $\alpha$ exists.}
\end{table}
	
\section{Real data application}
\label{sec:real}

\subsection{Multiple brain regions of rat}
\label{subsec:real1}

In the first example, we apply our method to an RNA-seq dataset studying the brain tissues of HIV transgenic rat from Gene Expression Omnibus (GEO) database \citep{li2013transcriptome}. RNA samples from three brain regions (hippocampus, striatum and prefrontal cortex) are sequenced for both control strains and HIV infected strains. Only the 36 control strains (12 samples in each brain region) are used here to see whether samples from the three brain regions can be correctly identified ($K=3, n_1=n_2=n_3=12$). After standard preprocessing and filtering out genes with mean counts smaller than 10 based on the guidance in edgeR \citep{robinson2010edger}, 10,280 genes are retained for clustering analysis. In this application, the true cluster labels (brain regions) are known and ARI can be used to evaluate clustering accuracy. However, the true informative genes are unknown and the AUC cannot be calculated to assess feature selection accuracy, as in simulation. Instead, we obtain results of a sequential number of selected genes (around 50-1000) by varying the tuning parameter and compare the ARI curves. Finally, we perform pathway enrichment analysis by using Fisher's exact test based on the Gene Ontology (GO), KEGG and Reactome pathway databases to assess the biological relevance of selected genes.




\begin{figure}
\centering
\centerline{\includegraphics[width=7in]{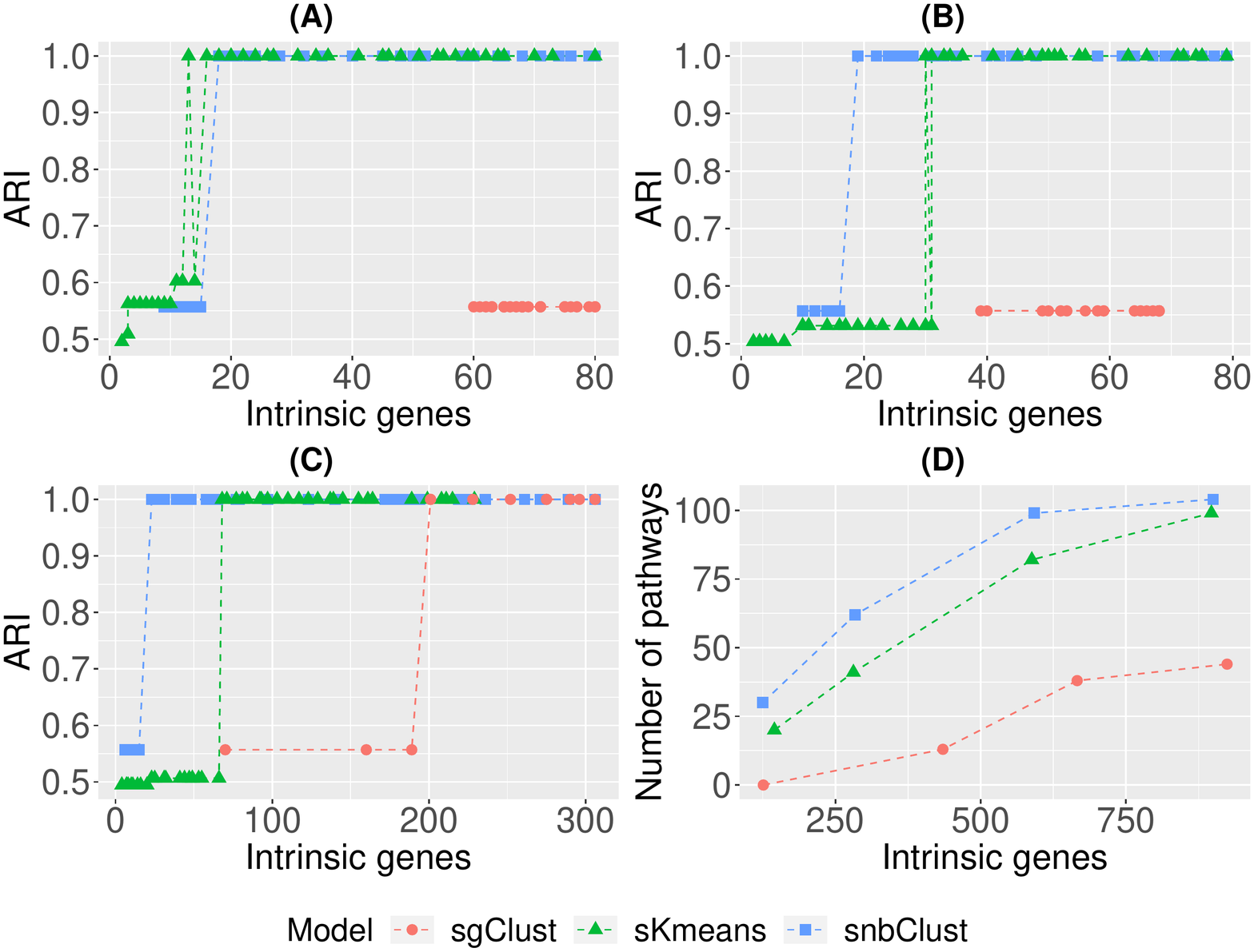}}
\caption{Figure \ref{f:3}(A) shows the ARI values under different gene selection for each method; Figure \ref{f:3}(B)(C) shows the clustering performance under 50\% and 20\% downsampling respectively; Figure \ref{f:3}(D) shows the number of enriched pathways under $FDR=0.05$ when different numbers of genes (by tuning $\lambda$) are selected.}
	\label{f:3}
\end{figure}
 Figure \ref{f:3}(A) shows the ARI values under different gene selection for each method. Both snbClust and sgClust demonstrate a perfect clustering performance (ARI=1) when more than 20 genes are selected while sgClust performs poorly with at most around ARI=0.55. To distinguish performance of snbClust and sgClust further, we randomly subsample the sequencing counts to mimic shallower sequencing experiments, which is commonly encountered to save experimental cost. Figure \ref{f:3}(B) and \ref{f:3}(C) show the ARI results when we downsampled the sequencing reads to only 50\% and 20\% of their original total reads. At 50\% subsampling sKmeans requires more than 30 selected genes to achieve perfect ARI and snbClust only needs 20. When sequencing depth is further reduced to 20\%, sKmeans needs 70 genes to achieve ARI=1 and snbClust only needs around 30 genes. The performance for sgClust has been found to be universally worse than the other two methods. When we used BIC or gap statistics to select the tuning parameter, sgClust and sKmeans select 9,846 and 10,280 genes respectively. The BIC of snbClust selects a more reasonable 1,311 gene set for clustering.
 
 To examine biological interpretation and functional annotation of selected genes, Figure \ref{f:3}(D) shows the number of enriched pathways under FDR=0.05 when different numbers of genes (by tuning $\lambda$) are selected. Compared to sKmeans and sgClust methods, snbClust consistenly detects more enriched pathways, implying the better functional association of selected genes by snbClust. Web Table 1 shows the union of pathways detected at FDR=0.01 using the top 284, 435 and 281 selected genes (35 for snbClust, 3 for sgClust and 29 for sKmeans; see Venn diagram in Web Figure 4(A)).  The result finds many neural development, synapse function and metal ion transport pathways that are known to be differentially active in different brain regions. 
 


\subsection{Breast Cancer dataset}
\label{sec:real2}
\begin{figure}
  \centerline{\includegraphics[width=7in]{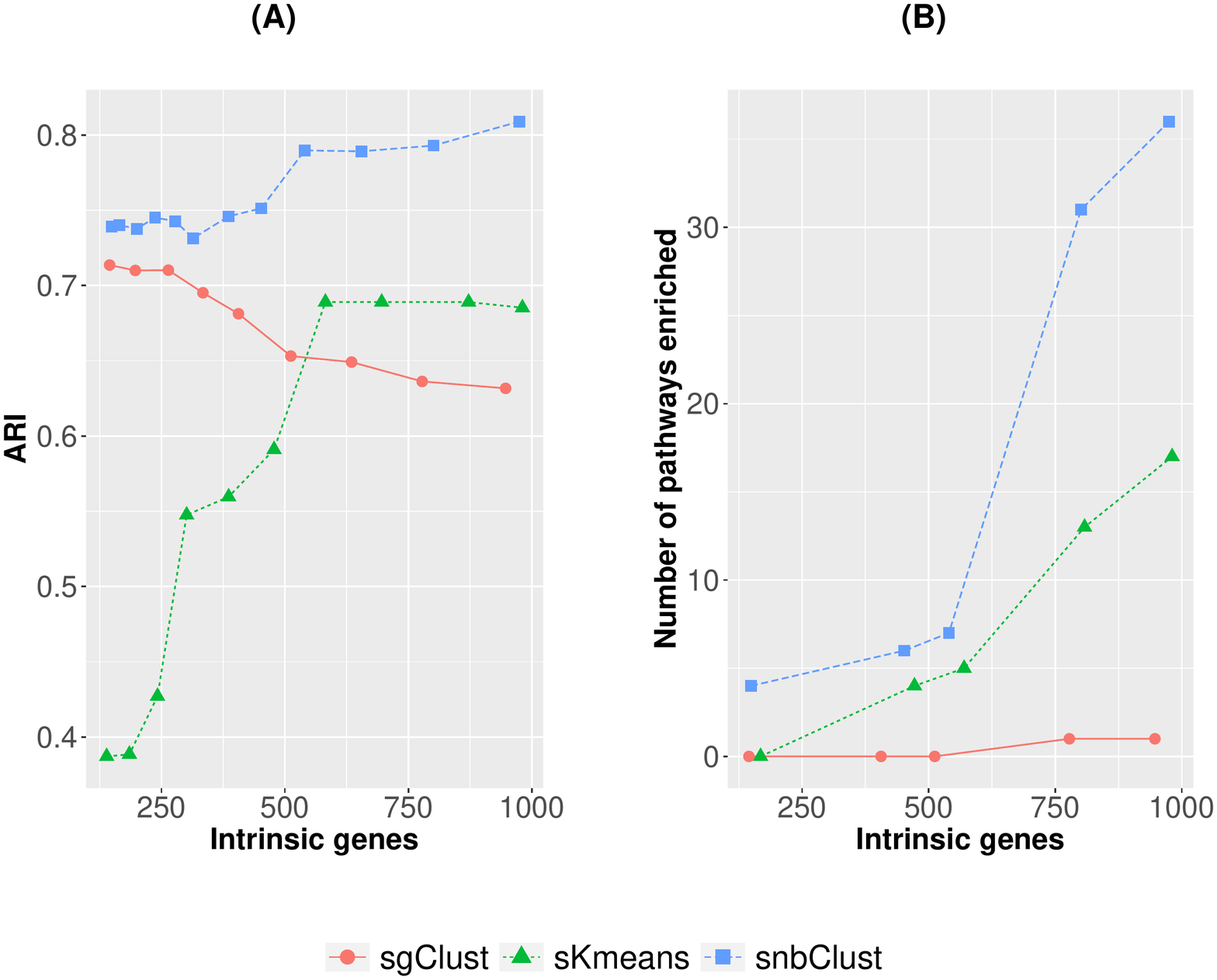}}
\caption{Comparison of snbClust, skmeans and sgClust model in Breast cancer data}
 \label{f:4}
\end{figure}

Next, we apply the three methods to the Cancer Genome Atlas (TCGA) breast cancer dataset. The dataset contains 610 female patients with four different subtypes of breast cancer: Basal (116 subjects), Her2 (63 subjects), LumA (257 subjects) and LumB (174 subjects). After standard preprocessing and using the criteria of filtering out genes with mean count less than 5 and variance less than the median variance, 8,789 genes are retained. LumA and LumB expression patterns were known to be similar, hence, three clusters considered for evaluation are Basal, Her2 and LumA+LumB. The evaluation is performed similarly to the rat brain example. As shown in Figure \ref{f:4}(A), snbClust reaches high clustering accuracy at 79\% when 540 genes are selected and outperforms sgClust and sKmeans (accuracy at most $\sim$70\%). In particular, performance of sgClust drops dramatically when the number of selected genes increases. In terms of pathway analysis, snbClust also performs the best with larger number of enriched pathways compared to the other two methods when selecting 127$\sim$1,000 top genes(Figure \ref{f:4}(B)). Specifically, we select similar numbers of selected genes (975, 947 and 981) for snbClust, sgClust and sKmeans and identify 33, 11 and 7 enriched pathways at FDR=0.05. Web Figure 4(B) shows the Venn diagram of enriched pathways from each method and Web Table 2 outlines the union set of 43 pathways. Web Table 2 contains many pathways known related to cancer. For example, cell-cell adhesion \citep{farahani2014cell}, ion channels \citep{biasiotta2016ion}, calcium signaling \citep{cui2017targeting}, transmembrane transporter activity \citep{huang2006membrane} and ectoderm development are related to tumorigenesis. Epidermis development is related to the HER2-enriched subtype, which requires specific treatment such as Trastuzumab \citep{iqbal2014human}. 
 \section{Discussion and conclusion}
\label{sec:discuss}
In this paper, we proposed a sparse model-based clustering analysis with negative binomial mixture distribution (snbClust). Since RNA-seq data are known to be discrete and skewed, negative binomial is a more appropriate distribution to capture the data characteristics, while normalizing counts to continuous and applying Gaussian-based models (e.g. sgClust and sKmeans) lose information and efficiency. The extensive simulations and two real applications clearly confirm this intuition, where snbClust outperforms sgClust and sKmeans in terms of clustering accuracy, gene selection and biological interpretation by pathway enrichment analysis. 

There are two potential limitations in the current model. Firstly, the new count data model requires heavier computing than Gaussian-based models although still in an affordable range for general omics applications. To benchmark computing time, sixteen choices of tuning parameter $\lambda$ are performed in simulation 2 using 1 computing core and average computing time for snbClust is comparable to sgClust (1.37 minutes versus 1.13 minutes). Besides, thirty choices of tuning parameter $\lambda$ are performed in breast cancer data using 10 computing cores. snbClust and sgClust take about 55 minutes and 42 minutes respectively. Similar to all optimization-based clustering algorithms, initial value plays an important role for successful clustering of all three methods. Secondly, the new model does not consider gene correlation structure that may be prevalent among the genes \citep{pen2}. Since the high dimensional data have number of features considerably larger than the number of samples and due to the complex structure of multivariate negative binomial distribution, incorporating correlation structure in the current model is not addressed in this paper and will be a future direction. We, however, have performed sensitivity analysis to examine performance impacted by existence of varying level of correlation structure. We find generally robust results in the clustering and feature selection using the current model with gene independence assumption. An R package is available on https://github.com/YujiaLi1994/snbClust, along with all data and source code used in this paper .


\section*{Acknowledgements}
Tanbin Rahman and Yujia Li contribute equally to this paper. YL and GCT are supported by NIH R01CA190766 and R21LM012752.

\bibliographystyle{biom} 
\bibliography{ref}

\begin{thebibliography}{}

\bibitem[\protect\citeauthoryear{Biasiotta, D'Arcangelo, Passarelli, Nicodemi,
  and Facchiano}{Biasiotta et~al.}{2016}]{biasiotta2016ion}
Biasiotta, A., D'Arcangelo, D., Passarelli, F., Nicodemi, E.~M., and Facchiano,
  A. (2016).
\newblock Ion channels expression and function are strongly modified in solid
  tumors and vascular malformations.
\newblock {\em Journal of translational medicine} {\bf 14,} 285.

\bibitem[\protect\citeauthoryear{Bouveyron and Brunet-Saumard}{Bouveyron and
  Brunet-Saumard}{2014}]{bouveyron2014model}
Bouveyron, C. and Brunet-Saumard, C. (2014).
\newblock Model-based clustering of high-dimensional data: A review.
\newblock {\em Computational Statistics \& Data Analysis} {\bf 71,} 52--78.

\bibitem[\protect\citeauthoryear{Cui, Merritt, Fu, and Pan}{Cui
  et~al.}{2017}]{cui2017targeting}
Cui, C., Merritt, R., Fu, L., and Pan, Z. (2017).
\newblock Targeting calcium signaling in cancer therapy.
\newblock {\em Acta pharmaceutica sinica B} {\bf 7,} 3--17.

\bibitem[\protect\citeauthoryear{Dempster, Laird, and Rubin}{Dempster
  et~al.}{1977}]{dempster1977maximum}
Dempster, A.~P., Laird, N.~M., and Rubin, D.~B. (1977).
\newblock Maximum likelihood from incomplete data via the em algorithm.
\newblock {\em Journal of the Royal Statistical Society: Series B
  (Methodological)} {\bf 39,} 1--22.

\bibitem[\protect\citeauthoryear{Donoho et~al\mbox{.}}{Donoho
  et~al.}{2000}]{donoho2000high}
Donoho, D.~L. et~al. (2000).
\newblock High-dimensional data analysis: The curses and blessings of
  dimensionality.
\newblock {\em AMS math challenges lecture} {\bf 1,} 32.

\bibitem[\protect\citeauthoryear{Eisen, Spellman, Brown, and Botstein}{Eisen
  et~al.}{1998}]{eisen1998cluster}
Eisen, M.~B., Spellman, P.~T., Brown, P.~O., and Botstein, D. (1998).
\newblock Cluster analysis and display of genome-wide expression patterns.
\newblock {\em Proceedings of the National Academy of Sciences} {\bf 95,}
  14863--14868.

\bibitem[\protect\citeauthoryear{Farahani, Patra, Jangamreddy, Rashedi,
  Kawalec, Rao~Pariti, Batakis, and Wiechec}{Farahani
  et~al.}{2014}]{farahani2014cell}
Farahani, E., Patra, H.~K., Jangamreddy, J.~R., Rashedi, I., Kawalec, M.,
  Rao~Pariti, R.~K., Batakis, P., and Wiechec, E. (2014).
\newblock Cell adhesion molecules and their relation to (cancer) cell stemness.
\newblock {\em Carcinogenesis} {\bf 35,} 747--759.

\bibitem[\protect\citeauthoryear{Fop, Murphy, et~al\mbox{.}}{Fop
  et~al.}{2018}]{fop2018variable}
Fop, M., Murphy, T.~B., et~al. (2018).
\newblock Variable selection methods for model-based clustering.
\newblock {\em Statistics Surveys} {\bf 12,} 18--65.

\bibitem[\protect\citeauthoryear{Fraley and Raftery}{Fraley and
  Raftery}{2002}]{fraley2002model}
Fraley, C. and Raftery, A.~E. (2002).
\newblock Model-based clustering, discriminant analysis, and density
  estimation.
\newblock {\em Journal of the American statistical Association} {\bf 97,}
  611--631.

\bibitem[\protect\citeauthoryear{Friedman, Hastie, and Tibshirani}{Friedman
  et~al.}{2010}]{friedman2010regularization}
Friedman, J., Hastie, T., and Tibshirani, R. (2010).
\newblock Regularization paths for generalized linear models via coordinate
  descent.
\newblock {\em Journal of statistical software} {\bf 33,} 1.

\bibitem[\protect\citeauthoryear{Huang and Sad{\'e}e}{Huang and
  Sad{\'e}e}{2006}]{huang2006membrane}
Huang, Y. and Sad{\'e}e, W. (2006).
\newblock Membrane transporters and channels in chemoresistance and-sensitivity
  of tumor cells.
\newblock {\em Cancer letters} {\bf 239,} 168--182.

\bibitem[\protect\citeauthoryear{Hubert and Arabie}{Hubert and
  Arabie}{1985}]{hubert1985comparing}
Hubert, L. and Arabie, P. (1985).
\newblock Comparing partitions.
\newblock {\em Journal of classification} {\bf 2,} 193--218.

\bibitem[\protect\citeauthoryear{Iqbal and Iqbal}{Iqbal and
  Iqbal}{2014}]{iqbal2014human}
Iqbal, N. and Iqbal, N. (2014).
\newblock Human epidermal growth factor receptor 2 (her2) in cancers:
  overexpression and therapeutic implications.
\newblock {\em Molecular biology international} {\bf 2014,}.

\bibitem[\protect\citeauthoryear{Kohonen}{Kohonen}{1998}]{kohonen1998self}
Kohonen, T. (1998).
\newblock The self-organizing map.
\newblock {\em Neurocomputing} {\bf 21,} 1--6.

\bibitem[\protect\citeauthoryear{Li, Cao, Wang, Wang, Sarkar, Vigorito, Ma, and
  Chang}{Li et~al.}{2013}]{li2013transcriptome}
Li, M.~D., Cao, J., Wang, S., Wang, J., Sarkar, S., Vigorito, M., Ma, J.~Z.,
  and Chang, S.~L. (2013).
\newblock Transcriptome sequencing of gene expression in the brain of the hiv-1
  transgenic rat.
\newblock {\em PLoS One} {\bf 8,} e59582.

\bibitem[\protect\citeauthoryear{Ma, Huang, and Zhang}{Ma
  et~al.}{2016}]{ma2016exploration}
Ma, S., Huang, J., and Zhang, Z. (2016).
\newblock Exploration of heterogeneous treatment effects via concave fusion.
\newblock {\em arXiv preprint arXiv:1607.03717} .

\bibitem[\protect\citeauthoryear{MacQueen et~al\mbox{.}}{MacQueen
  et~al.}{1967}]{macqueen1967some}
MacQueen, J. et~al. (1967).
\newblock Some methods for classification and analysis of multivariate
  observations.
\newblock In {\em Proceedings of the fifth Berkeley symposium on mathematical
  statistics and probability}, volume~1, pages 281--297. Oakland, CA, USA.

\bibitem[\protect\citeauthoryear{McLachlan}{McLachlan}{1997}]{mclachlan1997algorithm}
McLachlan, G. (1997).
\newblock On the em algorithm for overdispersed count data.
\newblock {\em Statistical Methods in Medical Research} {\bf 6,} 76--98.

\bibitem[\protect\citeauthoryear{Pan and Shen}{Pan and
  Shen}{2007}]{pan2007penalized}
Pan, W. and Shen, X. (2007).
\newblock Penalized model-based clustering with application to variable
  selection.
\newblock {\em Journal of Machine Learning Research} {\bf 8,} 1145--1164.

\bibitem[\protect\citeauthoryear{Robinson, McCarthy, and Smyth}{Robinson
  et~al.}{2010}]{robinson2010edger}
Robinson, M.~D., McCarthy, D.~J., and Smyth, G.~K. (2010).
\newblock edger: a bioconductor package for differential expression analysis of
  digital gene expression data.
\newblock {\em Bioinformatics} {\bf 26,} 139--140.

\bibitem[\protect\citeauthoryear{Schwarz et~al\mbox{.}}{Schwarz
  et~al.}{1978}]{schwarz1978estimating}
Schwarz, G. et~al. (1978).
\newblock Estimating the dimension of a model.
\newblock {\em The annals of statistics} {\bf 6,} 461--464.

\bibitem[\protect\citeauthoryear{Si, Liu, Li, and Brutnell}{Si
  et~al.}{2013}]{si2013model}
Si, Y., Liu, P., Li, P., and Brutnell, T.~P. (2013).
\newblock Model-based clustering for rna-seq data.
\newblock {\em Bioinformatics} {\bf 30,} 197--205.

\bibitem[\protect\citeauthoryear{Sugar and James}{Sugar and
  James}{2003}]{sugar2003finding}
Sugar, C.~A. and James, G.~M. (2003).
\newblock Finding the number of clusters in a dataset: An information-theoretic
  approach.
\newblock {\em Journal of the American Statistical Association} {\bf 98,}
  750--763.

\bibitem[\protect\citeauthoryear{Thalamuthu, Mukhopadhyay, Zheng, and
  Tseng}{Thalamuthu et~al.}{2006}]{thalamuthu2006evaluation}
Thalamuthu, A., Mukhopadhyay, I., Zheng, X., and Tseng, G.~C. (2006).
\newblock Evaluation and comparison of gene clustering methods in microarray
  analysis.
\newblock {\em Bioinformatics} {\bf 22,} 2405--2412.

\bibitem[\protect\citeauthoryear{Tibshirani and Walther}{Tibshirani and
  Walther}{2005}]{tibshirani2005cluster}
Tibshirani, R. and Walther, G. (2005).
\newblock Cluster validation by prediction strength.
\newblock {\em Journal of Computational and Graphical Statistics} {\bf 14,}
  511--528.

\bibitem[\protect\citeauthoryear{Tibshirani, Walther, and Hastie}{Tibshirani
  et~al.}{2001}]{tibshirani2001estimating}
Tibshirani, R., Walther, G., and Hastie, T. (2001).
\newblock Estimating the number of clusters in a data set via the gap
  statistic.
\newblock {\em Journal of the Royal Statistical Society: Series B (Statistical
  Methodology)} {\bf 63,} 411--423.

\bibitem[\protect\citeauthoryear{Tseng}{Tseng}{2007}]{tseng2007penalized}
Tseng, G.~C. (2007).
\newblock Penalized and weighted k-means for clustering with scattered objects
  and prior information in high-throughput biological data.
\newblock {\em Bioinformatics} {\bf 23,} 2247--2255.

\bibitem[\protect\citeauthoryear{Wang, Ma, Zappitelli, Parikh, Wang, and
  Devarajan}{Wang et~al.}{2016}]{wang2016penalized}
Wang, Z., Ma, S., Zappitelli, M., Parikh, C., Wang, C.-Y., and Devarajan, P.
  (2016).
\newblock Penalized count data regression with application to hospital stay
  after pediatric cardiac surgery.
\newblock {\em Statistical methods in medical research} {\bf 25,} 2685--2703.

\bibitem[\protect\citeauthoryear{Witten et~al\mbox{.}}{Witten
  et~al.}{2011}]{witten2011classification}
Witten, D.~M. et~al. (2011).
\newblock Classification and clustering of sequencing data using a poisson
  model.
\newblock {\em The Annals of Applied Statistics} {\bf 5,} 2493--2518.

\bibitem[\protect\citeauthoryear{Witten and Tibshirani}{Witten and
  Tibshirani}{2010}]{witten2010framework}
Witten, D.~M. and Tibshirani, R. (2010).
\newblock A framework for feature selection in clustering.
\newblock {\em Journal of the American Statistical Association} {\bf 105,}
  713--726.

\bibitem[\protect\citeauthoryear{Zhang et~al\mbox{.}}{Zhang
  et~al.}{2010}]{zhang2010nearly}
Zhang, C.-H. et~al. (2010).
\newblock Nearly unbiased variable selection under minimax concave penalty.
\newblock {\em The Annals of statistics} {\bf 38,} 894--942.

\bibitem[\protect\citeauthoryear{Zhou and Shen}{Zhou and Shen}{2009}]{pen2}
Zhou, H. and Shen, X. (2009).
\newblock Penalized model-based clustering with unconstrained covariance
  matrices.
\newblock {\em Electron J Stat.} {\bf 3,} 1473--1496.

\end{thebibliography}

\section*{Supporting Information}
Web Appendices, Tables, and Figures referenced are available with this paper at the Biometrics website on Wiley Online Library. The proposed method has been implemented in R. 





%


\end{document}